\documentclass[lettersize,journal]{IEEEtran}
\usepackage{amsmath,amsfonts}
\usepackage{algorithmic}
\usepackage{algorithm}
\usepackage{array}
\usepackage[caption=false,font=normalsize,labelfont=sf,textfont=sf]{subfig}
\usepackage{textcomp}
\usepackage{stfloats}
\usepackage{url}
\usepackage{verbatim}
\usepackage{multirow}
\usepackage{graphicx}
\usepackage{cite}
\usepackage{booktabs}
\usepackage{orcidlink}
\begin{document}

\title{Holistic Evaluation Metrics: Use Case Sensitive Evaluation Metrics for Federated Learning}

\author{Yanli Li\IEEEauthorrefmark{1}\orcidlink{0000-0001-9842-2209}, Jehad Ibrahim\IEEEauthorrefmark{1}\orcidlink{0009-0005-1256-1010
}, Huaming Chen\IEEEauthorrefmark{2}\orcidlink{0000-0001-5678-472X},~\IEEEmembership{Member,~IEEE,}, Dong Yuan\orcidlink{0000-0003-1130-0888} and Kim-Kwang Raymond Choo\IEEEauthorrefmark{2}\orcidlink{0000-0001-9208-5336}
\thanks{Yanli Li, Jehad Ibrahim, Huaming Chen and Dong Yuan are with the School of Electrical and Computer Engineering, The University of Sydney (e-mail: \{yanli.li, huaming.chen, dong.yuan\}@sydney.edu.au, jibr7656@uni.sydney.edu.au).}
\thanks{Kim-Kwang Raymond Choo is with the Department of Information Systems and Cyber Security, University of Texas at San Antonio, San Antonio, TX 78249, USA. (e-mail:raymond.choo@fulbrightmail.org).}
\thanks{\IEEEauthorrefmark{1}Equal contribution.}
\thanks{\IEEEauthorrefmark{2}Corresponding author.}
}

\maketitle
\begin{abstract}
A large number of federated learning (FL) algorithms have been proposed for different applications and from varying perspectives. However, the evaluation of such approaches often relies on a single metric (e.g., accuracy). Such a practice fails to account for the unique demands and diverse requirements of different use cases. Thus, how to comprehensively evaluate an FL algorithm and determine the most suitable candidate for a designated use case remains an open question. To mitigate this research gap, we introduce the Holistic Evaluation Metrics (HEM) for FL in this work. Specifically, we collectively focus on three primary use cases, which are Internet of Things (IoT), smart devices, and institutions. The evaluation metric encompasses various aspects including accuracy, convergence, computational efficiency, fairness, and personalization. We then assign a respective importance vector for each use case, reflecting their distinct performance requirements and priorities. The HEM index is finally generated by integrating these metric components with their respective importance vectors. Through evaluating different FL algorithms in these three prevalent use cases, our experimental results demonstrate that HEM can effectively assess and identify the FL algorithms best suited to particular scenarios. We anticipate this work sheds light on the evaluation process for pragmatic FL algorithms in real-world applications. 
\end{abstract}


\begin{IEEEkeywords}
Federated Learning (FL), Decentralized Machine Learning (DML), Evaluation Metrics, Collaboration Application
\end{IEEEkeywords}

\section{Introduction}
Federated learning (FL) is a trending distributed learning paradigm that enables many clients to train a machine learning model collaboratively while keeping the training data decentralized \cite{wen2023survey}. The underpinning privacy protection nature partly resulted in its application in various real-world scenarios, including \textit{smartphones} (smart devices \cite{gufran2023fedhil, wang2023applications}), \textit{institutional} settings \cite{guo2021multi}, and \textit{Internet of Things} (IoT) environment \cite{sheller2020federated}. For instance, Google implements the intelligent keyboard “Gboard” on its Pixel smartphones \cite{yang2018applied}, and Apple has developed its virtual assistant “Siri” for smart i-devices \cite{hao2020apple}. Studies such as \cite{rieke2020future, sheller2020federated} have explored the deployment of FL in health and medical institutions for achieving intelligent diagnosis and treatment. Furthermore, the research in \cite{bian2023client} demonstrates FL's significant potential in IoT scenarios, including vehicular traffic planning, unmanned aerial vehicles, and the smart city.

Different FL algorithms have been developed for various real-world applications \cite{chai2020fedeval} with different requirements, ranging from reduced communication costs (e.g., FedAvg \cite{kairouz2021advances}) to non-independent identical distribution (non-IID) settings (e.g., FedDyn \cite{Acar2021FederatedRegularization}, SCAFFOLD \cite{Karimireddy2019SCAFFOLD:Learning}) to and personalized local model (e.g., MAML, ProtoNet \cite{FallahPersonalizedApproach}), etc. Given the diversity of FL algorithms, it is no surprise that selecting an appropriate candidate for a specific use case is an important research trend \cite{kairouz2021advances}. To inform the selection, existing studies typically assess and compare FL algorithms based on a single metric, such as accuracy or loss value. While such evaluation indexes can partially reflect the performance of FL algorithms from particular perspectives, they fall short in providing a comprehensive assessment of the overall model performance \cite{kairouz2021advances}. Furthermore, the evaluations based on a single metric fail to consider the unique requirements and priorities of different FL use cases, further undermining the reliability of the evaluation outcome. To mitigate the research gap, we introduce holistic evaluation metrics (HEM) and a corresponding evaluation pipeline to thoroughly assess FL algorithms and select the most appropriate FL candidate for the targeted use cases. 

Given the unique performance requirements of each FL use case, we start by highlighting the three most representative contexts, namely smartphones (smart devices), institutions, and IoT \cite{li2020review, mills2021multi}. Our proposed evaluation framework, the HEM index, goes beyond a singular metric by encompasses multiple key components, including Client Accuracy, Convergence, Computational Efficiency, Fairness, and Personalization (for personalized FL-PFL algorithms), ensuring a comprehensive assessment. The HEM for any FL algorithm is generated by a weighted average of these component performance indices, where the weights assigned are proportional to the particular needs identified (referred to as importance vectors) for the target application. Specifically, we identify that accuracy and fairness are the primary performance requirements for FL applications in IoT and institutional contexts. In contrast, convergence and computational efficiency are the main focus in smartphone related applications. 

In this study, we evaluate a range of FL algorithms and personalization methods across these identified cases using our proposed HEM and the corresponding evaluation pipeline. The experimental results demonstrate that the HEM can effectively evaluate and differentiate FL algorithms, aiding in the algorithm selection for target use cases. While we currently focuses on three FL use cases in this paper, our HEM could be easily extended to other application scenarios through further investigation into the importance of different metrics components.

The main contributions of our work are: 
\begin{itemize}
\item We establish a comprehensive set of metrics, including accuracy, convergence, computational efficiency, fairness, and personalization for organizing the evaluation metrics.
\item We identify the specific needs of different use cases and assess the importance of each corresponding component accordingly.
\item We propose the HEM index based on the evaluation components and importance identified, achieving comprehensive and effective evaluation for an FL algorithm in the designated use cases.
\end{itemize}

This work has substantially extended our previous study~\cite{jehad2024holistic}. 
Specifically, we extend Section \ref{IIA} for a more comprehensive survey on the FL application, Section \ref{IIID} about the detailed HEM evaluation process, and Section \ref{IVD} with the insight of the performance trade-off by introducing personalization.    

The rest of this paper is organized as follows. The next section provides related works surveying existing federated learning applications, algorithms, and evaluation metrics. Section \ref{section:Holistic Evaluation Metrics} describes the proposed HEM, followed by the experimental results and related discussions in Section \ref{DataResults}. The last section concludes the paper.

\section{Related Works}\label{RW}
\subsection{Federated Learning Applications}\label{IIA}
To utilize data isolated at client locations without compromising privacy, numerous applications have been deployed employing FL architecture. These applications typically fall into three categories, depending on the use cases. They include applications for smart phones (smart devices) \cite{yang2018applied, tabassum2023depression, varlamis2023using}, institutions \cite{ogier2023federated, lo2022toward}, and IoT \cite{wang2021electricity, zhang2022federated, wu2022fedadapt}. 

Firstly, Google's successful deployment of its FL-based user input prediction keyboard (Gboard) has sparked significant interest in developing FL applications for smart devices. This application category of smart devices now includes not only text prediction of user input \cite{yang2018applied} but also emoji prediction \cite{ramaswamy2019federated}, monitoring health condition (such as depression detection) \cite{tabassum2023depression}, and offering energy efficiency plan recommendations to users \cite{varlamis2023using}. Secondly, FL's capabilities in preserving privacy have led to its adoption in data-sensitive institutional settings. A recent study \cite{ogier2023federated} introduces a para-medicine application designed to predict the histological response to neoadjuvant chemotherapy (NACT) in early-stage breast cancer patients. FL has facilitated the utilization of clinical information while ensuring data is securely maintained behind hospital firewalls. To support the treatment for patients with COVID-19, another study \cite{dayan2021federated} proposes an FL-based prediction model, named EXAM, which successfully uses the data from 20 institutes across the globe without compromising privacy. Thirdly, FL is being applied in the IoT sector to address the limitations of high storage costs, together with privacy concerns associated with the traditional ecosystem of centralized over-the-cloud learning and IoT platforms. To offer diversified electricity services, the study \cite{wang2021electricity} introduces an FL application that identifies electricity consumer characteristics. Specifically, privacy-preserving principal component analysis (PCA) is utilized to extract features from smart meter data, while an artificial neural network is trained in a federated setting with various weighted averaging strategies.

\subsection{Federated Learning Algorithms}
FL allows multiple participants to collaboratively train a shared model without sharing their raw data \cite{kairouz2021advances}. Typically, a FL training round includes three steps: the server first broadcasts the current global model to all participants. Subsequently, each participant locally updates the global model using their own private data, sending the updated model back to the server once the established training criteria are satisfied. Finally, the server aggregates the models received using a designated FL algorithm and finalizes the current training round \cite{mcmahan2017communication, Konecny2016FederatedEfficiency}.

As a key FL algorithm, FedAvg \cite{mcmahan2017communication} efficiently aggregates the client gradients while minimising the communication costs. In the FedAvg approach, the server computes a weighted average of the clients' models, where the weights are proportional to the size of the training data each client possesses. Despite reducing the communication bottleneck in FL, FedAvg stuggles with handling non-IID data and heterogeneous data, limiting its deployment in real-world scenarios. To address these shortcomings and further improve FL performance, FedDyn \cite{Acar2021FederatedRegularization} and SCAFFOLD \cite{Karimireddy2019SCAFFOLD:Learning} have been developed. The algorithm FedDyn builds on the concept of adding a penalty term to the objective function introduced by FedProx but dynamically changes the value of the penalty term by using a novel dynamic regularization method to ensure that the minima of the global and local objective functions are consistent \cite{Acar2021FederatedRegularization}. SCAFFOLD is an algorithm that builds on FedAvg by attempting to improve its convergence limitation when trained on non-IID data. It corrects the “client drift” issue by leveraging variant reduction. 

In addition to ensuring identical model functions for all participants, there is a growing research focus on developing personalized FL algorithms and providing customised learning performances. The study \cite{Alp2021DebiasingTraining} proposes two algorithms, PFLDyn and PFLScaf, which apply the gradient correction methods (FedDyn and SCAFFOLD) to correct bias in a meta-model and tailor personalised client models. Two meta-learning approaches are used to implement the algorithms: MAML and ProtoNet \cite{Alp2021DebiasingTraining}. The algorithms present convergence guarantees for convex and nonconvex meta objectives. Further, the research \cite{FallahPersonalizedApproach} proposes a personalized version of the FedAvg algorithm, namely Per-FedAvg, which trains an initial shared model using the MAML framework. The experimental findings indicate that the solution derived from Per-FedAvg can facilitate a more personalized outcome.

\begin{table*}[h]
    \centering
    \begin{tabular}{l|c|c|c|c}
        \toprule
        \midrule
        \multirow{2}*{\textbf{FL/PFL Algorithms}} & \multicolumn{4}{c}{\textbf{Evaluation Criteria}} \\ 
        \cline{2-5}

        ~& \textbf{Accuracy} & \textbf{Convergence} & \textbf{Complexity} & \textbf{Fairness} \\
        \midrule
        FedAvg & \checkmark & \checkmark & $\times$ & $\times$ \\ \midrule
        FedAvg (MAML) & $\times$ & \checkmark & $\times$ & $\times$ \\ \midrule
        FedAvg (Proto) & \checkmark & \checkmark & $\times$ & $\times$ \\ \midrule
        FedDyn & \checkmark & \checkmark & \checkmark & $\times$ \\ \midrule
        FedDyn (MAML) & \checkmark & \checkmark & $\times$ & $\times$ \\ \midrule
        FedDyn (Proto) & \checkmark & \checkmark & $\times$ & $\times$ \\ \midrule
        SCAFFOLD & \checkmark & \checkmark & \checkmark & $\times$ \\ \midrule
        SCAFFOLD (MAML) & \checkmark & \checkmark & $\times$ & $\times$ \\ \midrule
        SCAFFOLD (Proto) & \checkmark & \checkmark & $\times$ & $\times$ \\
        \midrule
        \toprule
    \end{tabular}
    \caption{Criteria used for different FL algorithms evaluations.}
    \label{tab:algorithm-evaluation}
\end{table*}

\subsection{Federated Learning Evaluation Metrics}
Currently, the evaluation metrics used in FL generally derive from those established in traditional centralized machine learning (CML) concepts. The development of specific metrics tailored for FL is still in its nascent stages. Commonly, metrics like accuracy or loss value, which measure the percentage of correct prediction on the testing data or quantify the difference between predictions and actual results, are used to evaluate both FL and CML algorithms \cite{Alp2021DebiasingTraining, shayan2020biscotti}. In addition, since FL training relies on regular communication between the server and clients, some research evaluates algorithm performance based on communication efficiency \cite{imteaj2021survey}, considering the factors such as the number of communication rounds, parameters, and message sizes required to train an FL model \cite{Alp2021DebiasingTraining}. From a system performance perspective, the evaluation of an FL algorithm could be conducted based on its network and hardware requirements \cite{li2020review, kairouz2021advances}. Furthermore, several studies \cite{kairouz2021advances, li2020review} employ metrics that assess local model fairness and robustness against adversarial attacks, thereby assessing the trustworthiness of their frameworks.

We broadly review the key FL algorithms (i.e., FedAvg, FedDyn, and SCAFFOLD) and delve into two meta-learning techniques (i.e., MAML and Proto) that can adapt these FL algorithms into personalized FL (PFL) algorithms. We note that, although these algorithms could effectively address specific weaknesses of the original algorithm, the researchers used to evaluate and demonstrate improvements using a single metric. This approach often make the trade-offs involved with each algorithm unclear. 

Thus, in this work, we propose the Holistic Evaluation Metric (HEM), aiming to provide an effective and comprehensive evaluation of FL algorithms and aid in selecting the most appropriate FL algorithm for diverse real-world use cases. Table \ref{tab:algorithm-evaluation} illustrates the criteria used for different FL algorithm evaluations.

\section{Holistic Evaluation Metrics} \label{section:Holistic Evaluation Metrics}
In this section, we present Holistic Evaluation Metrics (HEM) to mitigate the research gap carried by current simplistic evaluation metric and provide an effective evaluation method for FL algorithms in various real-world use cases. Recognising that different use cases may prioritize different performance aspects, we start by identifying the most representative FL use cases. We then determine the principal evaluation perspectives, and finally organize the HEM by developing the formula for the HEM index.

\subsection{Federated Learning Use Cases}
FL has revolutionzed traditional ML by facilitating training on decentralized data while addressing crucial privacy concerns in privacy-sensitive applications where training data is distributed across multiple edge devices \cite{ye2023heterogeneous}. Currently, a large number of service providers have adopted the FL method, which has found a significant role in different domains, ranging from smart devices to institutions and IoT. Herein, we explore the three representative real-world application use cases of FL, drawing on insights from the applications and characteristics outlined in recent studies \cite{pandya2023federated, wen2023survey}.

\subsubsection{Smartphones (Smart Devices)}
Smart devices, including smartphones and laptops, significantly benefit humans and have become integral to our daily lives. While providing information to users, these devices also collect user information and usage patterns. Applications such as next-word prediction, face detection, and voice recognition \cite{Hard2018FederatedPrediction, yang2021characterizing} could greatly benefit from these collected and recorded data from users. However, data gathering in a centralized location to support traditional centralized ML poses challenges due to user privacy concerns and legal restrictions. FL addresses these issues by facilitating the collaborative learning of ML models across a wide array of mobile devices. The smartphone participants can retain the data locally, only sharing the model update with the server via the cellular or wireless network. By leveraging FL, on the one hand, the vast amounts of user data ensure both high learning performance and satisfactory service provision; on the other hand, clients benefit from personalized functions through their participation. For instance, providers of next-word predictor applications can employ FL to train model with the historical text data from users across a broad network of smart devices. In return, clients benefit from more accurate prediction results that are customized to their individual user habits \cite{LiFederatedDirections}.

\subsubsection{Institutions}
Existing institutions maintain large amounts of records and information on their clients to deliver services \cite{sheller2020federated}. For example, healthcare institutions store diagnostic logs of patients to inform treatment strategies, while financial institutions keep records of clients' bank statements to make lending decisions. While such data could greatly improve training datasets and boost model performance, these institutions are frequently subject to strict privacy regulations and ethical considerations, necessitating that data remains localized \cite{Joshi2022FederatedChallenges}. This limitation hinders institutions' ability to utilize data from other organizations or even from their client data, ultimately constraining their potential to offer enhanced intelligent services. FL presents an innovative solution for these use cases by ensuring privacy and enabling institutions to participate in secure collaborations. Within an FL system, institutions acts as “client” nodes sharing only the model updates during each learning iteration, thus maintaining data privacy within their boundaries \cite{Joshi2022FederatedChallenges}. Furthermore, vertical FL technology allows the deployment of an FL model in scenarios where participants share the same sample ID space but possess different feature sets, effectively accommodating collaboration between institutions even from different domains.
    
\subsubsection{Internet of Things (IoT)}
The IoT ecosystem consists of a network of interconnected devices, such as wearable devices\cite{chen2020fedhealth}, autonomous vehicles \cite{li2021privacy}, and smart homes \cite{nguyen2022federated}, all equipped with sensors to collect, process, and act on real-time data. For example, autonomous vehicles rely on continually updated models for safe navigation, taking into account traffic conditions, construction sites, and pedestrian movements. However, building accurate and up-to-date models can be challenging due to the privacy concerns associated with data and the limited connectivity of each device \cite{wang2022ai}. Moreover, the large volume of data generated by IoT devices every second further brings difficulty to the traditional centralized ML. To this end, FL algorithms are considered a viable alternative, which adapt IoT devices to dynamic environments, enabling efficient model training while preserving user privacy. When deploying FL in an IoT use case, each interconnected device can act as a client contributing partially or fully to the model training, depending on its computing resources \cite{zhang2021optimizing}. A trustworthy cloud node serves as the server, orchestrating the entire training process.

\begin{table*} [h]
\setlength{\tabcolsep}{12pt}
    \centering
    \begin{tabular}{c|c|c|c} \hline 
        \toprule[1pt]
        \midrule
         \textbf{Use Cases}& \textbf{IoT}& \textbf{Smart devices (Smartphone)}& \textbf{Institution}\\ \midrule
         Accuracy& High & Moderate & High\\ \midrule
         Convergence& Low & High & Low\\ \midrule 
         Computational Efficiency& High& High & Low\\ \midrule
         Fairness& High&Moderate&High\\ 
         \midrule
         \toprule[1pt]
    \end{tabular}
    \caption{Holistic evaluation metric component importance for IoT, smartphone, and institution.}
    \label{t1}
\end{table*}

\subsection{Evaluation Metric Components}
Rather than relying solely on a single performance metric, such as accuracy or loss value, we organize the HEM through multiple components to provide a more comprehensive evaluation. Specifically, these evaluation metric components encompass client accuracy, convergence, computation efficiency, fairness, and personalization for PFL.

\textbf{Evaluation Metric Component 1:  Client Accuracy}\\ 
Accuracy, as the most intuitive indicator, is widely used in the evaluation of existing ML and FL algorithms to assess model learning performance. Within the FL context, the global model used to be evaluated using testing data after each iteration's aggregation.  The percentage of accurate predictions is then calculated as accuracy, serving as a measure to demonstrate the overall learning performance. However, due to the heterogeneity of participants' data and devices, their learning performance can vary significantly. Therefore, in the proposed HEM, we leverage client accuracy over global model accuracy, calculating it as the mean accuracy across clients per round. Client Accuracy reflects the utility and learning performance of the FL algorithm for individual clients under the given learning task and data distribution. Specifically, a higher Client Accuracy indicates greater satisfaction among clients regarding the predictions generated by their models. It essentially provides a clear measure of the FL algorithm's effectiveness and ability to meet the varied needs and expectations of participating clients.

\textbf{Evaluation Metric Component 2: Convergence}\\ 
The FL training heavily relies on the regular models (or gradients) exchange between clients and the server to perform the learning process. An algorithm that requires numerous learning rounds to achieve convergence may introduce high communication costs and exacerbate the bottleneck. To this end, we introduce the Convergence component in the HEM. The Convergence component reflects the time and computational resources an FL algorithm consumes by measuring the number of communication rounds an FL algorithm requires to achieve a predetermined target accuracy for clients. A higher convergence score indicates a heavier demand for communication computational and temporal resources, potentially taxing the devices involved. In our study, we set the target accuracy at 80\%, a benchmark deemed satisfactory for model performance across all investigated use cases. Additionally, we normalize the Convergence value on a scale from 0 to 1000, where 0 represents optimal performance and 1000 indicates the worst outcome.

\textbf{Evaluation Metric Component 3: Computation Efficiency}\\ 
While HEM utilizes the Convergence component to represent the time cost by counting the communication rounds, the training time per round can vary across different FL algorithms and finally drives impact on the learning efficiency. To this end, we further include Computational Efficiency in HEM to introduce clock time cost evaluation. Specifically, the Computational Efficiency component reflects the amount of time or memory required for a given threshold but emphasizes the wall-clock time. In the HEM, the Computational Efficiency is measured using a metric called Time to Accuracy (TTA). TTA merges the concepts of Convergence and Simulation Time, providing a view of the duration necessary for an algorithm to meet the set target accuracy or threshold. Besides, This component evaluates the balance between computational resource demand and time investment and demonstrates an algorithm's efficiency in resource utilization.

\textbf{Evaluation Metric Component 4: Fairness}\\
The Fairness reflects the level of the learning performance (model accuracy) difference that exists in the FL system across all participants. In HEM, the  Fairness component is assessed by calculating (inversely proportional to) the entropy ($H$) of the Client Accuracy list. Formally, for a given accuracy list $L={\{a_{1}, ..., a_{i}}\}, |L|=I$ , where $a_{i}$ denotes the accuracy of the $i$th client. The Fairness ($F$) of $L$ is:

\begin{equation}
    1/F(L)\propto H(L)=-\sum_{i=0}^{I}a_ilog_{e}a_{i}
\end{equation}

A lower entropy signifies a more equitable distribution of accuracy, while a higher entropy points to disparities in performance among clients. Due to the data and device heterogeneity, clients usually achieve different learning performances in FL systems. Introducing Fairness in the evaluation metric avoids overemphasizing the overall global model and ignores the participant's disadvantage. It illustrates the balance of accuracy among the clients, indicating the algorithm's ability to address the diverse capabilities and characteristics of participating devices, so-called fairness. 

\textbf{Evaluation Metric Component 5: Personalization (PFL)}\\
In FL systems, clients may have various exceptions for the global model due to their different interests and use patterns. To evaluate the effectiveness of customization achieved through personalization methods, we introduce the Personalization component into the HEM for PFL algorithms. This component is calculated by comparing the Client Accuracy list with and without introducing personalization methods. Within HEM, Personalization is represented as the median percentage improvement in accuracy across all clients when using the PFL algorithm compared to the original FL algorithm. A higher personalization index indicates the method's capability to enhance performance for individual clients' specific data distribution, demonstrating strong potential for delivering personalized functions in real-world applications.

\subsection{Holistic Evaluation Metrics (HEM)} \label{Importance Tables}
Now, we construct the HEM using the identified components of the evaluation metric. This subsection presents the organization of HEM across the three most representative use cases—IoT, smartphone, and institution; however, the HEM is versatile and can be applied to any other use case. 

The HEM is generated through a linear combination of each evaluation metric component value and represented as an index ranging from 0 to 1, with 1 denoting the ideal algorithm for a specific use case. Specifically, each evaluation metric component is assigned an importance level (namely Importance Vector) between 0 and 1, considering the specific requirements of the use case. Then, the holistic index is computed as a weighted average of the evaluation metric component indexes, taking into account the significance of each component within the given use case. 

Formally, the HEM is generated followed Equation. (2):
\begin{equation}
HEM=\sum_{i=1}^{N}Index_{i}\cdot \stackrel{UseC}{Impor V_{i}}\label{5}
\end{equation} 
where ${Index}_{i}$ and $ImporV_i$ indicate the value (Index) of each evaluation metric component and its associated Importance Vector, respectively. $UseC$ demonstrates the target use case, and $N$ indicates the amount of evaluation metric component. Due to the unique characteristic of each use case, the $\stackrel{UseC}{Impor V_{i}}$ should be different. TABLE.~\ref{t1} illustrates the HEM component importance for IoT, smartphone and institution use cases, and we will provide detailed information and discussion in the following subsections.

\begin{figure*} [h]
    \centering
    \includegraphics[width=\linewidth]{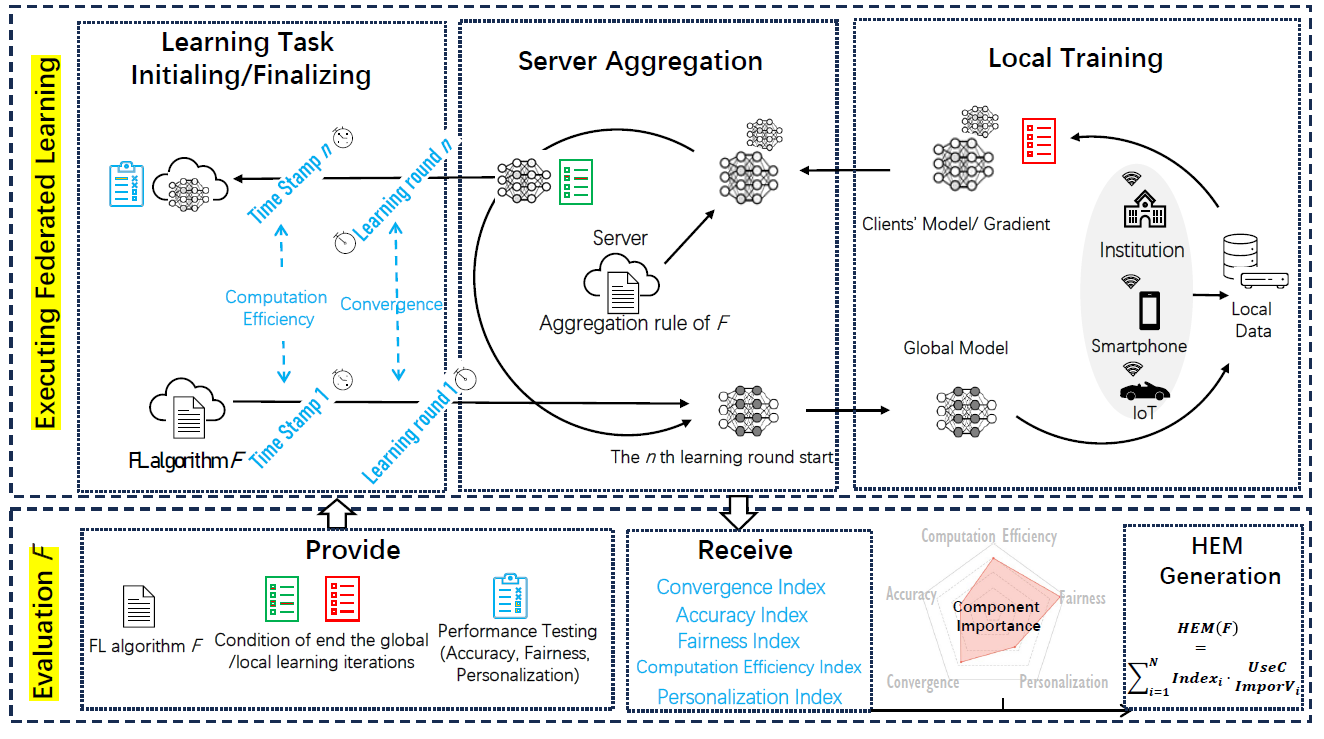}
    \caption{The overall evaluation process with HEM for PFL in different use cases.}
    \label{fig:process}
\end{figure*}

\subsubsection{HEM for IoT Use Cases} 
Given the advancements in FL and IoT technologies, heavy machinery (including autonomous vehicles and cranes) can now operate autonomously, significantly reducing the human workload \cite{imteaj2021survey, khan2021federated}. However, given their considerable volume, weight, and strength, any inaccuracies in the decisions made by these machines could lead to life-threatening outcomes. Therefore, maintaining a high level of accuracy for every single participant in IoT scenarios is crucial, justifying the High importance of the evaluation components - Client Accuracy and Fairness. Similarly, the Computational Efficiency component is also deemed as High important due to the limited computational resources of most IoT devices. We note that although some of these devices are allowed to train FL models in the background without impacting their primary functions, their reduced CPU power and memory capacity restrict the effectiveness of such background processes, which emphasizes the priority of Computational Efficiency. On the contrary, due to the IoT devices typically remaining idle for an extended duration, which could make the execution of numerous FL communication rounds \cite{mills2019communication}, the importance of the model Convergence is considered as Low level.

\subsubsection{HEM for Smartphone Use Cases}
Today, smartphones (as well as smart devices) provide numerous applications to the user based on their platform. To derive the importance of each metric component for the smartphone use case, we take the “next word prediction” as an example \cite{Hard2018FederatedPrediction}. Through literature review, we note that most next-word prediction applications do not reach very high accuracy (over 90\%) and only support general prediction services \cite{khurana2023natural, min2023recent}. On the other hand, while service providers aim to deliver customized predictions, creating a personalized model for each individual client within a network is impractical due to the vast number of users. Thus, in the use case of “next word prediction,” we rank the importance of Client Accuracy and Fairness as Moderate. As the environment changes dynamically with the user, smartphones usually suffer from fluctuating bandwidth and network. This unstable connection between client nodes and the server emphasizes the requirement of effectiveness Convergence of the FL model, with the importance level as High \cite{min2023recent}. Furthermore, as most smartphones are designed with limited computational resources to facilitate portability, we classify the importance of Computational Efficiency as High level.

\subsubsection{HEM for Institution Use Cases}
The institutional use cases include all organizations or institutions that are sensitive to data privacy and seek to benefit from FL systems.
Here, we take healthcare institutions \cite{Joshi2022FederatedChallenges} as the example to derive and investigate the importance level for each component of HEM. Given that medical diagnoses contain highly sensitive patient information, individuals may not be willing to participate in the FL task and contribute their data if the algorithm does not offer a satisfactory service. Thus, Client Accuracy and Fairness achieve a High importance level in institutional use cases. On the other hand, institutions usually have stable networking, high bandwidth, and sufficient computational resources. They can afford computing-intensive learning tasks and achieve regular communication between the server and end nodes. Thus, the Convergence and Computational Efficiency are rated as Low importance.\\

While we emphasize the importance of each component covering the most applications in three use cases, we acknowledge that certain specific applications within these categories may have unique requirements. Therefore, the importance levels indicated in Table \ref{t1} are intended as general guidelines, subject to further alignment based on the particular demands of individual applications.

\subsection{Evaluation Process} \label{IIID}
In this section, we discuss the evaluation process from preparation through to the generation of HEM. 

During the HEM evaluation procedure, the evaluator starts by selecting candidates FL algorithms, identifying the learning tasks that fit the specific use case, and establishing the termination criteria for local and global model training. These criteria include the target accuracy and a preset training round. The testing datasets should be collected to facilitate both global and local model testing, ensuring that the selection of testing data reflects the non-IID data distribution of FL and the given use case.

When performing the evaluation, each selected FL algorithm undergoes normal training within the specified scenario, during which both the training duration and the number of learning rounds are recorded until the predefined end conditions are satisfied. To derive all evaluation components, the delivered model under selected FL algorithms may be initialized and undergo multiple training times, concluding upon meeting various predefined thresholds. Following training, the models are assessed using testing datasets to generate index values from various perspectives. Considering the Importance Vector specific to the use case, these index values are then converted into HEM scores. While Table \ref{t1} provides a guideline about the components' importance in different use cases, these importance levels could be further updated based on the requirements of the given scenario and applications. These HEM scores finally facilitate the informed selection of FL algorithms for subsequent use. The evaluation process through HEM is shown in Figure \ref{fig:process}.

\section{FL Algorithms Evaluation through HEM} \label{DataResults}
In this section, we evaluate the most prominent FL algorithms and personalization methods using the proposed HEM index and the corresponding evaluation process. Initially, the evaluation focuses on each individual component of the metric. We then integrate these components into the HEM framework using the formula referenced in Equation (2). THis section concludes with a discussion of the findings.

\begin{table*} [h]
\small
    \centering
    \begin{tabular}{c|c|c|c|c}
        \toprule[1pt]
        \midrule
        \textbf{FL Algorithm} & \textbf{Client Accuracy} & \textbf{Convergence} & \textbf{Computation Efficiency} & \textbf{Fairness}\\
        \midrule
        FedAvg & 0.84  & 0.67 & 0.12 & \textbf{1.00}\\         
        \midrule
        FedAvg\_MAML & 0.88 & 0.90 & 0.00 & 0.55\\         
        \midrule
        FedAvg\_Proto & 0.88 & 0.85 & 0.78 & 0.36 \\         
        \midrule
        FedDyn & 0.85 & 0.69 & 0.21 & 0.41\\         
        \midrule
        FedDyn\_MAML & \textbf{0.89} & \textbf{0.94} & 0.56 & 0.00\\         
        \midrule
        FedDyn\_Proto & \textbf{0.89} & 0.92 & \textbf{0.89} & 0.27\\ 
        \midrule
        SCAFFOLD & 0.86 & 0.86 & 0.44 & 0.59\\         
        \midrule
        SCAFFOLD\_MAML & 0.87 & 0.86 & 0.67 & 0.23\\         
        \midrule
        SCAFFOLD\_Proto & \textbf{0.89} & 0.91 & 0.87 & 0.05\\
        \midrule
        \toprule[1pt]
    \end{tabular}
    \caption{The accuracy, convergence, computation efficiency, fairness index (performances) of different FL algorithms.}
    \label{tab:performance}
\end{table*}

\subsection{Experiments Setup} 
We consider 100 clients participating in the FL tasks; each client trains the model with data on only 5 out of 10 classes in the chosen dataset. We set the end conditions as “1000 communication rounds” to generate the Client Accuracy index and “80\% accuracy” to generate the Convergence index. We use The Cifar-10 as the training dataset, which consists of 60000, 32x32 pixels color images in 10 classes, including airplanes, cars, birds, cats, deer, dogs, frogs, horses, ships, and trucks. 

As the overall data distribution in FL is considered non-IID, we introduce a custom dataset configuration to tailor the dataset to the requirements of our FL simulations.
Specifically, we distribute 5 classes of training and test data to each end node based on its class list. The $i$th client is assigned the class list shown as follows, where $\%$ denotes operation $MOD$:
\begin{equation}
   (i+n) \% 10, n\in [0,1,2,3,4]
\end{equation}

We select the 5 most representative Fl algorithms for experiments, including FedAvg, FedDyn, SCAFFOLD, MAML (for personalized), and ProtoNet (for personalized). We use a CNN in our experiments, which has two convolutional layers (64× and 5×5 kernels), a max pooling layer, two fully connected layers (384×, 192×) with ReLU activation, and a softmax layer. 

\subsection{Evaluation Metric Component Indexes} \label{Component Indexes}
Table \ref{tab:performance} illustrates the Client Accuracy, Convergence, Computation Efficiency, and Fairness index (evaluation outcomes) of different FL algorithms and their personalized implementation.

\subsubsection{Client Accuracy Index}
The experimental results show that all FL algorithms achieve a high client model testing accuracy, with a variance in the accuracy index among them of less than 0.05.  FedDyn, utilizing both MAML and Proto personalization methods, and SCAFFOLD, with Proto personalization, attain the highest performance, each achieving an accuracy of 0.89. In contrast, FedAvg witnesses the lowest testing accuracy at 0.84. The accuracy gap can be attributed to FedAvg's performance degradation when facing heterogeneous and non-IID data \cite{Karimireddy2019SCAFFOLD:Learning, LiFederatedDirections}. Furthermore, personalization methods such as MAML and Proto allow individual clients to receive FL models that are better fitted to their unique data distributions. Consequently, algorithms personalized through these methods show approximately 3\% higher accuracy compared to their original algorithms.

\subsubsection{Convergence Index}
The Convergence Index across all simulated FL algorithms exhibits a wider range than the Client Accuracy, with the lowest performing algorithm, FedAvg, demonstrating a difference of approximately 0.17 when compared to the highest performing algorithm, FedDyn\_MAML. Specifically, FedDyn\_MAML, FedAvg\_MAML, FedDyn\_Proto, and SCAFFOLD\_Proto show a higher performance, achieving a Convergence Index of over 0.90. In contrast, the lowest Convergence Index is observed in FedAvg and FedDyn, both falling below 0.7. This convergence performance difference is attributed to the accuracy gains in PFL algorithms achieved through the effectiveness of meta-learning techniques. Such techniques enable FL algorithms to transfer knowledge across clients, facilitating faster convergence \cite{FallahPersonalizedApproach}. In contrast, non-personalized FL algorithms may not consistently converge in limited communication rounds, particularly when trained on clients with strong non-IID data distribution (as simulated).

\subsubsection{Computational Efficiency Index}
The Computational Efficiency index is a comparative index across the simulated algorithms. As achieving the longest TTA, FedAvg\_MAML is assigned to the lowest Computational Efficiency Index with 0. The experimental results indicate a strong correlation between the Computational Efficiency of FL algorithms and the personalization methods employed. Specifically, for each base FL algorithm, the incorporation of Proto personalization results in better computational efficiency compared to the integration of MAML. Besides, while recent computational efficiency research indicates FedAvg outperformed SCAFFOLD when the client datasets were IID \cite{Karimireddy2019SCAFFOLD:Learning}, we note SCAFFOLD heavily outperformed FedAvg in non-IID scenarios. For instance, FedAvg, MAML, and Proto achieve 0.12, 0.00, and 0.78 Computation Efficiency Index, while SCAFFOLD, MAML, and Proto achieve 0.44, 0.67, and 0.87, respectively.

\subsubsection{Fairness Index}
The Fairness index serves as a comparative measure across the simulated algorithms, scaling from 0 to 1. Among the three non-personalized FL algorithms, FedAvg achieves the highest Fairness Index (indicating the lowest entropy) with a score of 1.00. It is followed by SCAFFOLD with a score of 0.59, FedAvg\_MALA at 0.55, and FedDyn at 0.41. In contrast, FedDyn\_MAML exhibits the highest entropy in its accuracy list, resulting in the lowest Fairness index of 0. Besides, the experimental results indicate that incorporating personalization methods into these FL algorithms increases the variance in clients' accuracy and leads to a reduction in the Fairness Index. Specifically, FedAvg, FedDyn, and SCAFFOLD exhibit reductions in their Fairness Index ranging from 0.45 to 0.64, 0.14 to 0.41, and 0.36 to 0.54, respectively.

\subsubsection{Personalization of PFL Algorithms} \label{sec:PersonalizationResults}
The experimental results demonstrate that two personalization methods, Proto and MAML, effectively enhance model learning performance based on the client's specific data distribution, resulting in positive personalization indices for all evaluated algorithms. Specifically, FedAvg\_Proto emerged as the most personalized FL algorithm, achieving a Median Percentage of Client Accuracy Improvement (MPI) of 10.46, closely followed by FedAvg\_MAML. The FedDyn class of PFL algorithms demonstrates moderate levels of personalization, receiving MPI values of 9.2 and 8.11 for the Proto and MAML methods, respectively. Conversely, SCAFFOLD PFL algorithms show the lowest degree of personalization, with an average MPI value of around 8.

\begin{figure} [h]
    \centering
    \includegraphics[width=1.0\linewidth]{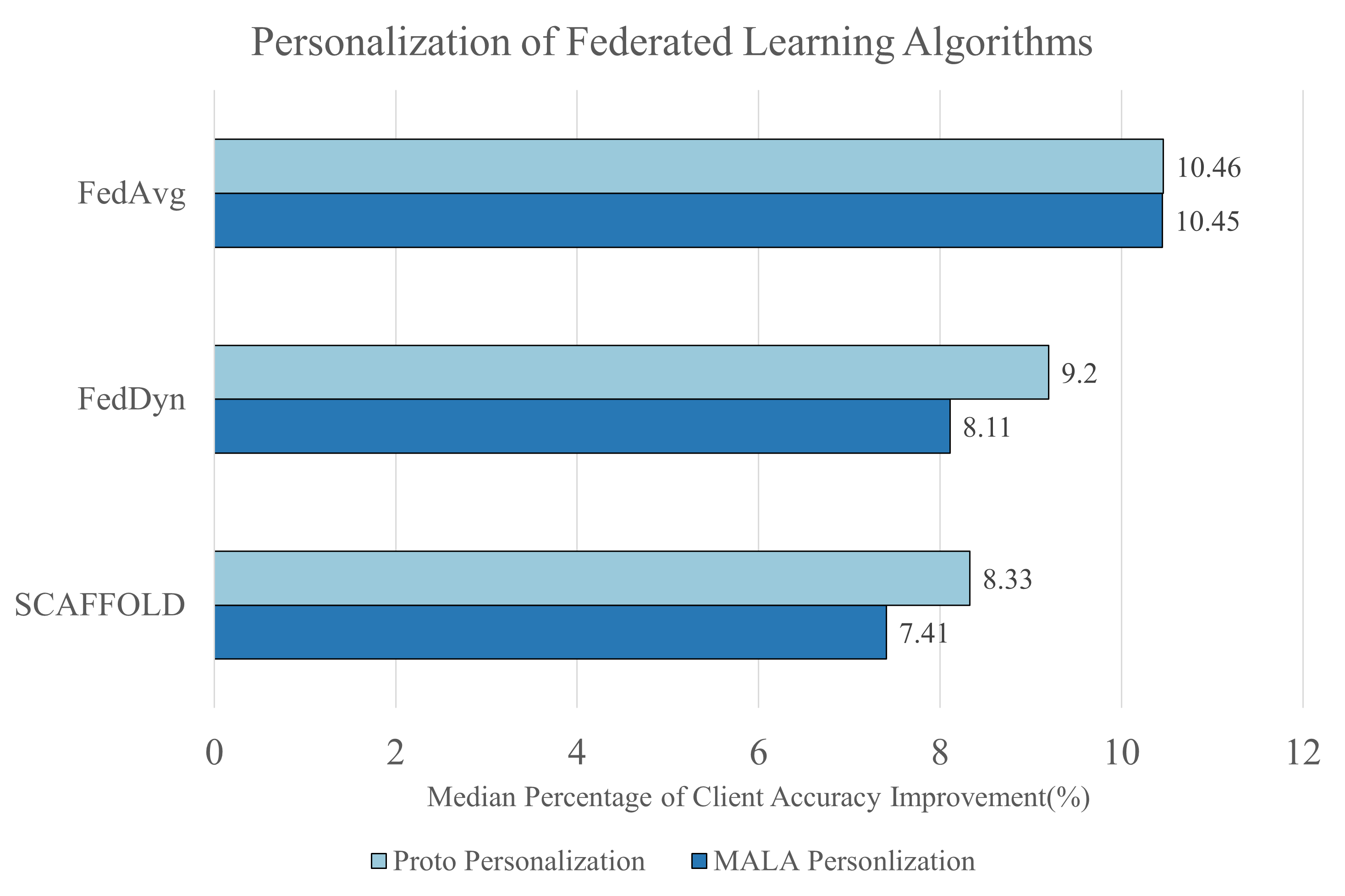}
    \caption{Personalization of PFL algorithms}
    \label{fig:Personalization}
\end{figure}

Figure \ref{fig:Personalization} illustrates the personalization capabilities of PFL algorithms.

\subsection{Holistic Evaluation for the Three Use Cases}
In this section, we conduct the HEM evaluation of the FL algorithms across the identified use cases (IoT, institution, and smartphone) based on the experimental results presented in Section \ref{Component Indexes} and Equation \ref{5}. The importance levels are quantified as follows: High is assigned an index of 3, Moderate is an index of 2, and Low is an index of 1 for organizing HEMs. Under this simulation, an FL algorithm exhibits Excellent overall performance if its HEM index is greater than 0.8. Performance is classified as Good for HEM indices ranging from 0.7 to 0.8, Acceptable for indices between 0.5 and 0.7, and Low for indices below 0.5. These quantification and standards set here could be further updated according to the special requirements of target use cases.  

\subsubsection{IoT Use Case} \label{section:IoTEval}

\begin{figure} [h]
    \centering
    \includegraphics[width=1.0\linewidth]{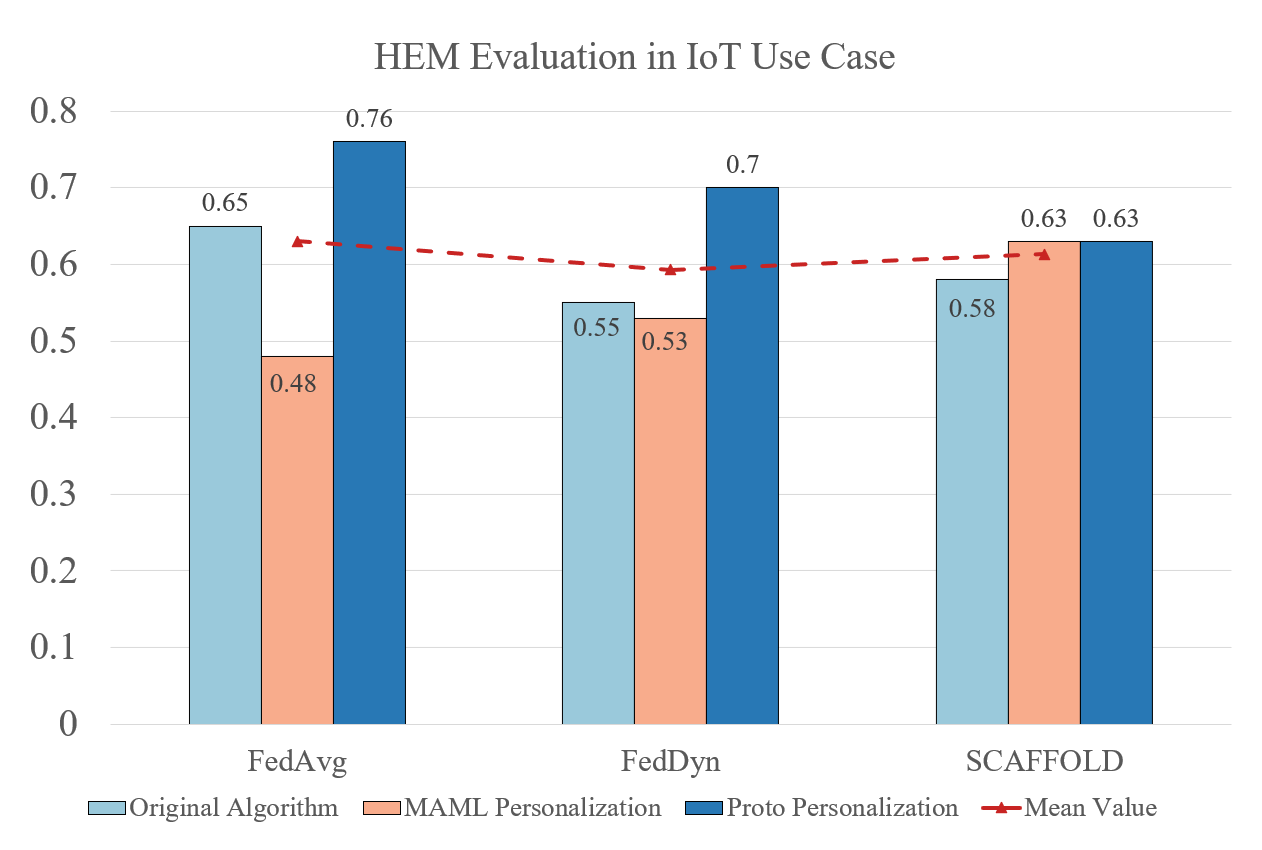}
    \caption{HEM index of various FL algorithms in IoT use case.}
    \label{fig:iot}
\end{figure}

In the IoT use case, the HEM index indicates that the FL algorithms' performance ranges from Good to Low. Specifically, FedAvg\_Proto and FedDyn\_Proto achieve the highest performance, with a HEM index of 0.76 and 0.70, respectively. FL algorithms based on SCAFFOLD receive the middle level with an average HEM index of around 0.60. In contrast, FedAvg\_MAML and FedDyn\_MAML receive the lowest HEM index, approximately 0.50. This variation in HEM performance is attributed to the High importance of Accuracy, Computational Efficiency, and Fairness in IoT use cases. Thus, FL algorithms that demonstrate high performance in these components (e.g., FedAvg\_Proto) maintain an advantage in the HEM evaluation, whereas those with lower scores in these aspects (e.g., FedAvg\_MAML) are at a disadvantage. Figure \ref{fig:iot} illustrates the HEM index of various FL algorithms in IoT use cases.

Recall the individual competent performance evaluated shown in Table \ref{tab:performance}, FedAvg\_Proto algorithm does not reach any highest performance on each single perspective. As a result, a user seeking an FL algorithm for IoT use cases through traditional evaluation metrics may overlook the FedAvg\_Proto, while it emerges as the most suitable algorithm when considering the IoT requirements. However, our proposed HEM reveals a different insight and shows the benefits of the FedAvg\_Proto algorithm in the target scenarios. This comprehensive evaluation underscores the importance of considering the collective strengths of FedAvg\_Proto rather than focusing on isolated metrics.

\subsubsection{Smartphone Use Case} \label{section:SmartphoneEval}

\begin{figure}[h]
    \centering
    \includegraphics[width=1.0\linewidth]{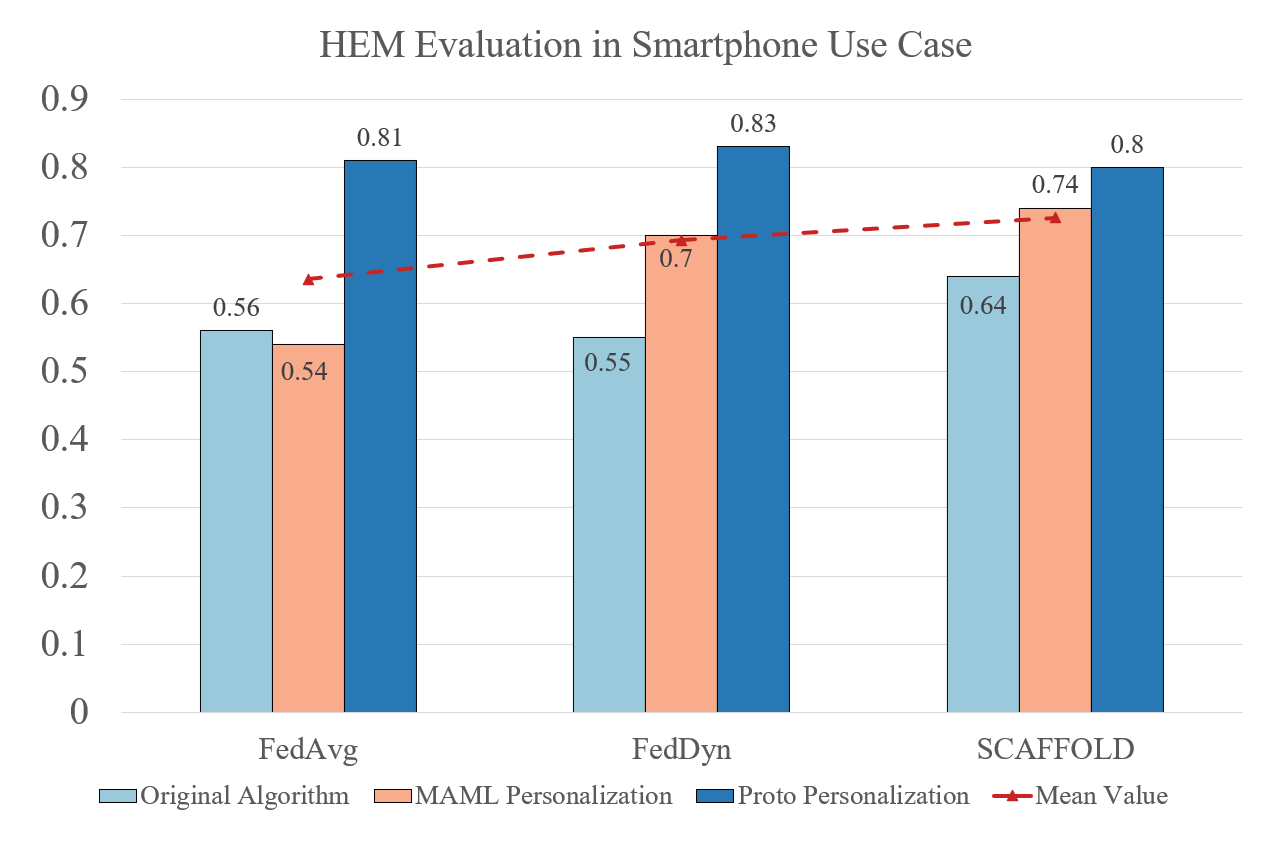}
    \caption{HEM index of various FL algorithms in the Smartphone use case.}
    \label{fig:smart}
\end{figure}

The HEM index shows that the performance of FL algorithms in the Smartphone use case ranges from Acceptable to Excellent. On the one hand, all three PFL algorithms that use the Proto meta-learning framework score in the Excellent Performance, achieving a HEM index of over 0.80. On the other hand, the original algorithms generally receive a low HEM index, with FedAvg at 0.56, FedDyn at 0.55, and SCAFFOLD at 0.64. From an average performance perspective, the SCAFFOLD class achieves the highest mean value at 0.72, followed by FedDyn at 0.69 and FedAvg at 0.63, indicating its advantages in Smartphone use cases. Figure \ref{fig:smart} illustrates the HEM index of various FL algorithms in the Smartphone use case.
\begin{figure}[h] 
    \centering
    \includegraphics[width=1.0\linewidth]{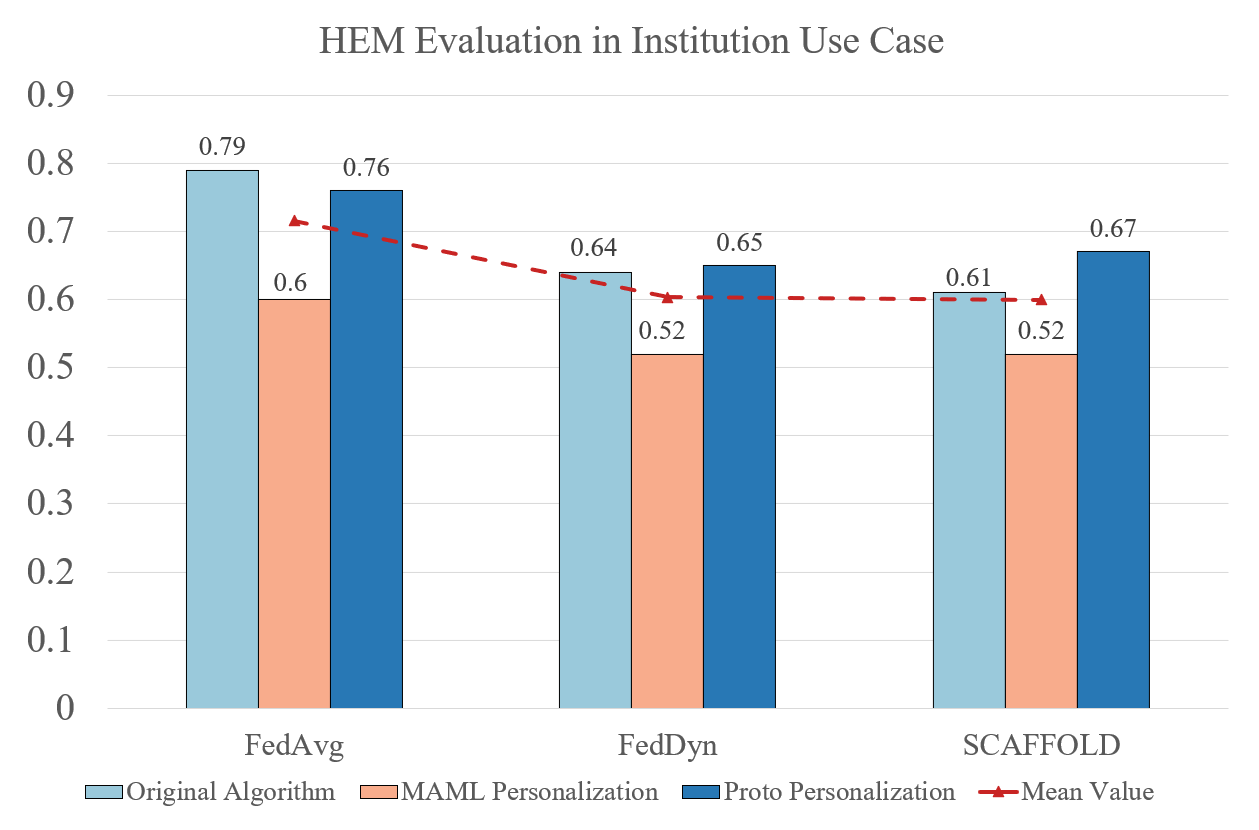}
    \caption{HEM index of various FL algorithms in Institution use case.}
    \label{fig:ins}
\end{figure}
Given that Convergence and Computational Efficiency are highly valued in the HEM evaluation for the smartphone use case, an FL algorithm that excels in these areas can compensate for deficiencies in other components. Thus, although FedDyn\_Proto may not consistently achieve the highest performance in existing evaluations, it could be identified through the HEM as the most suitable algorithm for Smartphone users.

\subsubsection{Institution Use Case} \label{InstitutionEval}

\begin{figure*}[h] 
    \centering
    \includegraphics[width=0.329\linewidth]{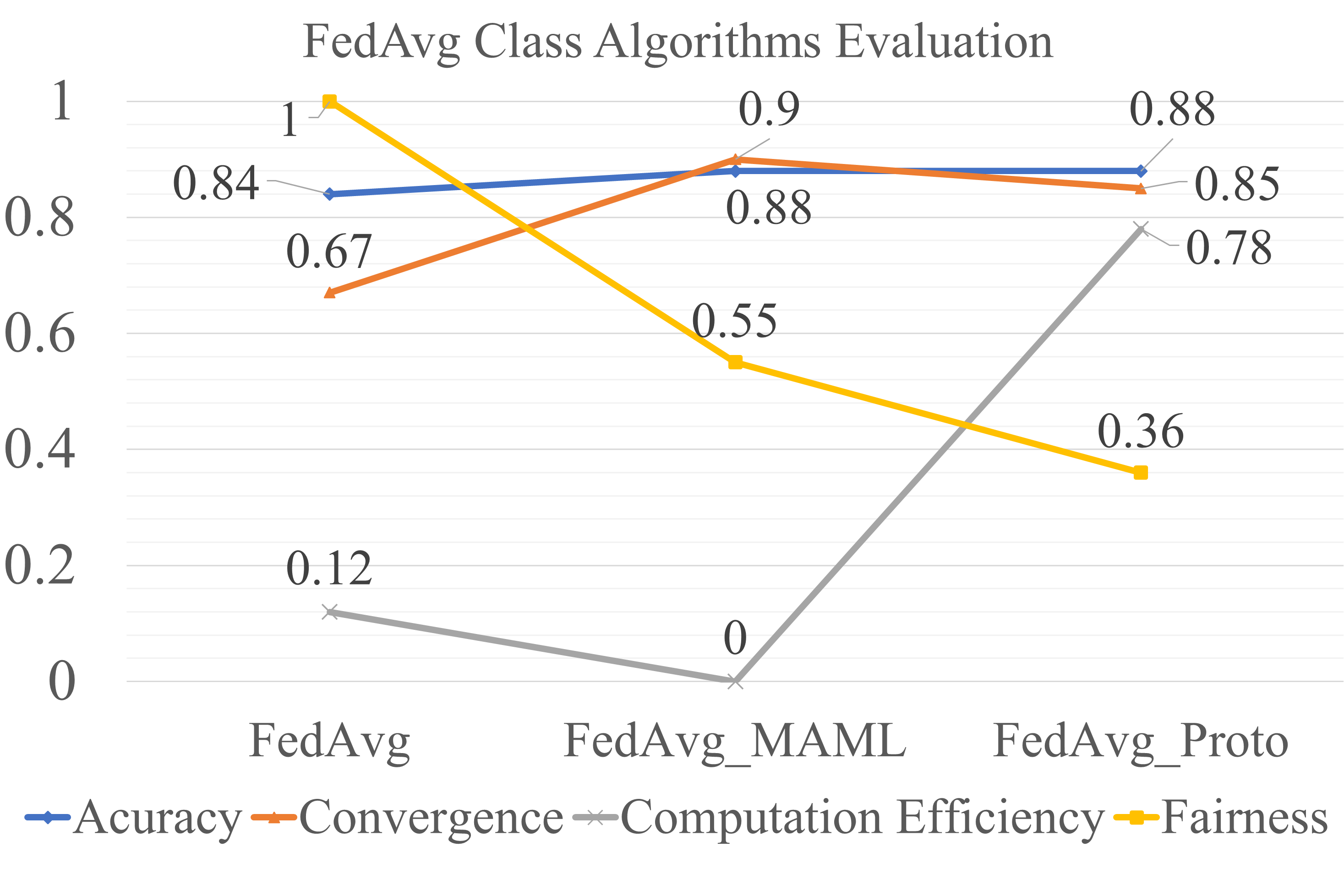}
    \includegraphics[width=0.329\linewidth]{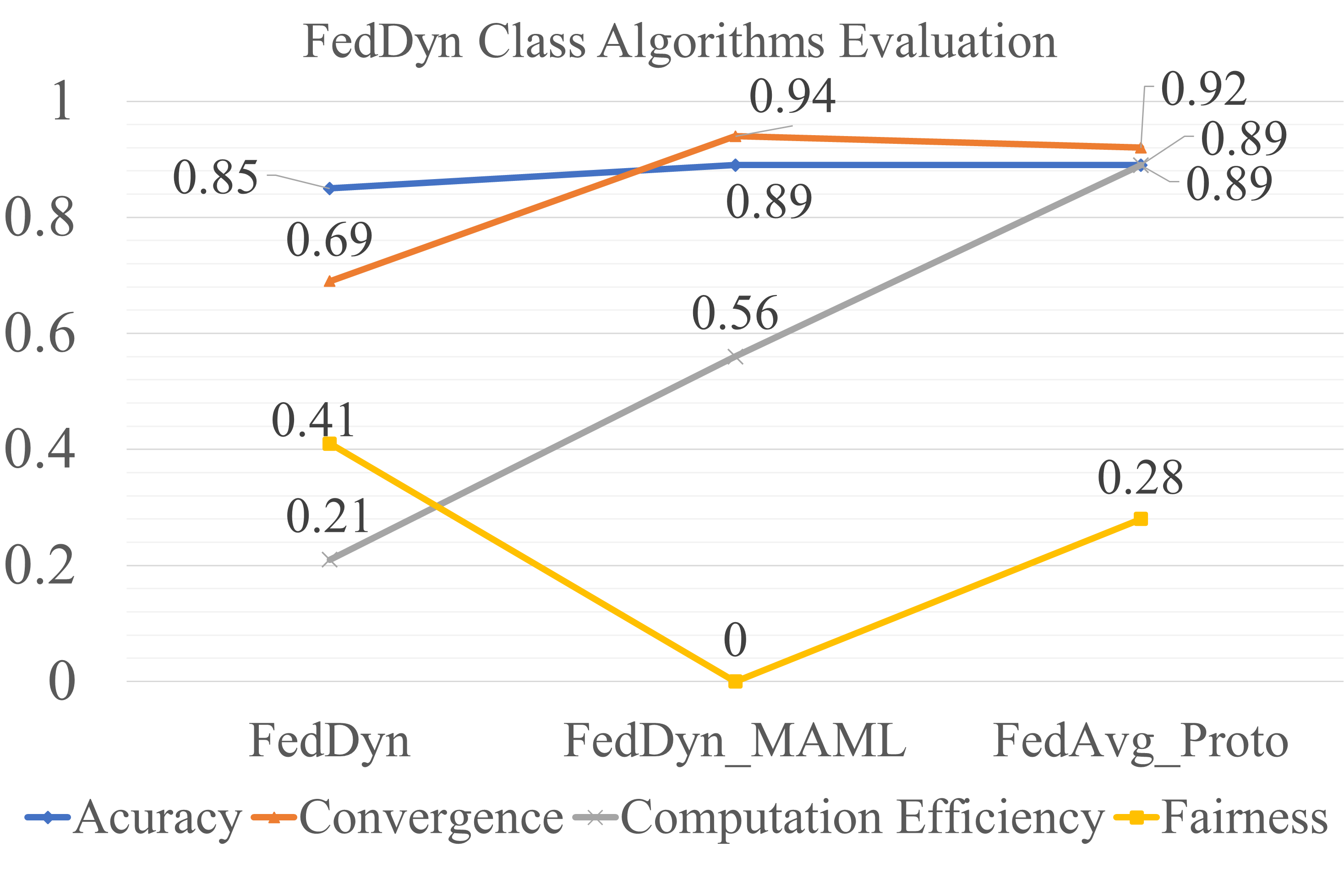}
    \includegraphics[width=0.329\linewidth]{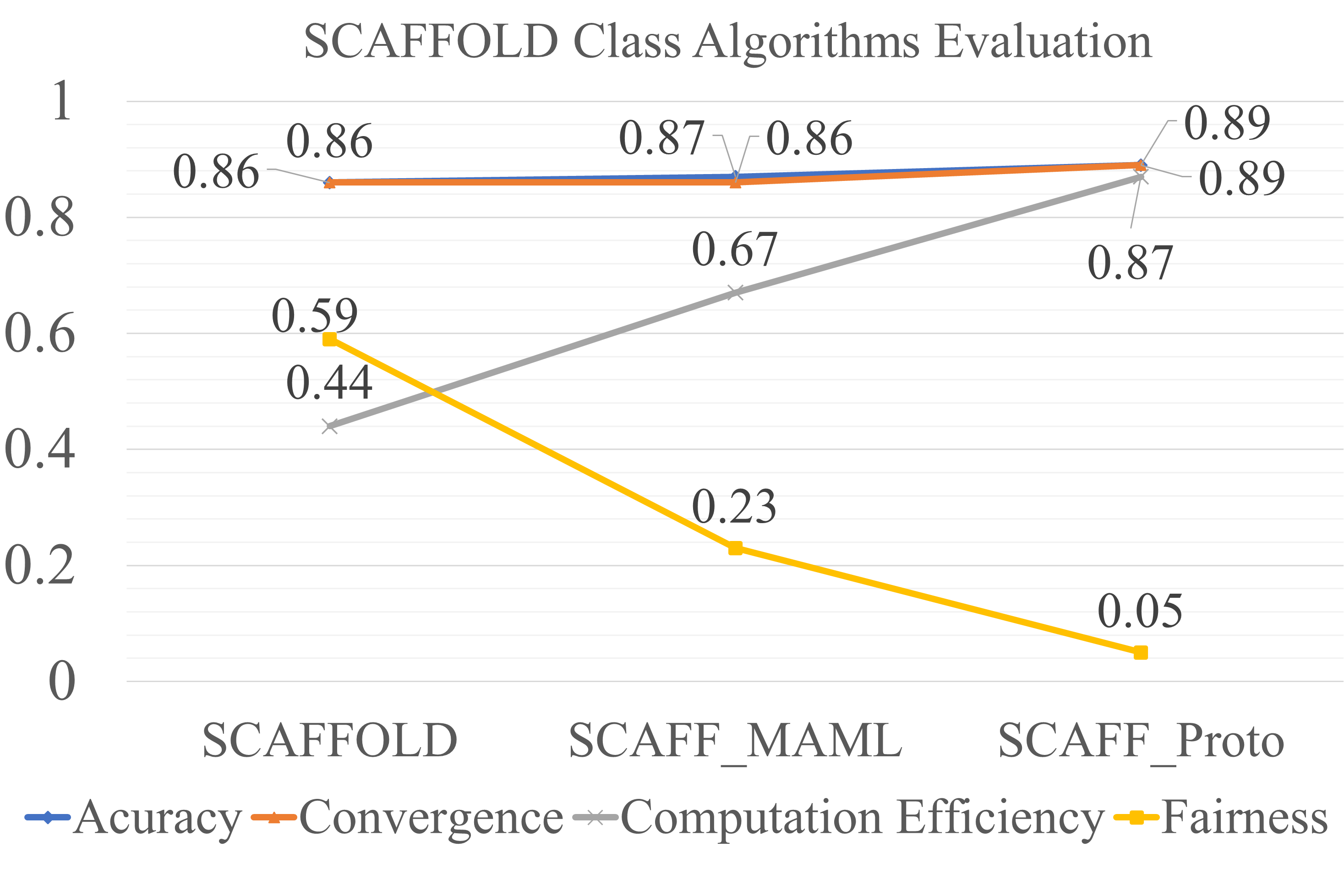}
    \caption{Illustration of the index fluctuation in various evaluation components upon introducing personalization methods to the FedDyn, SCAFFOLD, and FedAvg algorithms.}
    \label{dis}
\end{figure*}
In the institution use case, the algorithms show performance from Good to Acceptable. Specifically, FedAvg scores the highest with a HEM index of 0.79, followed by FedAvg\_Proto at 0.76 and FedDyn\_Proto at 0.67. In contrast, FedDyn\_MAML and SCAFFOLD achieve the lowest performance, which receives a 0.52 HEM index. As the Institution use cases prioritize Client Accuracy and Fairness, the HEM index for FedDyn\_MAML is adversely impacted by its lower performance in fairness despite achieving high model accuracy. Additionally, it is observed that most FL algorithms experience a decrease in their HEM index upon integrating MAML personalization, whereas the Proto method tends to maintain or even enhance performance. Figure \ref{fig:ins} illustrates the HEM index of various FL algorithms in the simulated institution use case.

Based on the HEM evaluation results, a user seeking an FL algorithm to support model training and delivery services in institutional use cases might consider selecting FedAvg\_Proto and FedDyn\_Proto. Both algorithms deliver Good overall performance and satisfactory Fairness, which could continuously encourage clients to participate in the learning task.

\subsection{Discussion} \label{IVD}

The previous experimental results demonstrate that personalization methods enhance the Client Accuracy and Computation Efficiency of original FL algorithms by effectively converging on heterogeneous or non-IID data during the training process (see Table \ref{tab:performance}). However, it remains unclear whether these improvement carried by personalization methods implicitly brings trade-offs that could negatively affect other components and the overall HEM. 

Thus, in this subsection, we seek to answer the following three research questions: 

\textbf{RQ1:} What kind of trade-offs are carried by personalization methods? 

\textbf{RQ2:} Whether the PFL algorithms with high personalization performed better?

\textbf{RQ3:} Whether the FL algorithms in each simulated use case are benefited from the implementation of personalization methods?

Figure \ref{dis} illustrates the index fluctuation of various evaluation components by introducing personalization methods (MAML, Proto) to the FedDyn, SCAFFOLD, and FedAvg algorithms. One can observe that almost all components experience an increase following the introduction of personalization methods. Specifically, through introducing personalization, the Computational Efficiency achieves the largest improvement, with gains of up to approximately 0.7. Convergence and Accuracy experience slighter improvements, approximately around 0.03 and 0.2, respectively. However, all Fairness indexes witnessed a significant decrease when the FL algorithm was personalized, with reductions of up to 0.64. We attribute this to the fact that PFL can achieve performance improvements for only a subset of clients and shows limited effectiveness for others with disadvantaged data distributions. The unbalance performance improvement across clients enlarges the accuracy difference and subsequently decreases the Fairness. Thus, while personalization methods can enhance the performance of FL algorithms, particularly in terms of Computational Efficiency, they sacrifice Fairness among participants, leading to significant accuracy disparities in non-IID scenarios. When introducing personalization methods, such a trade-off should be considered. A PFL method may be more suitable for use cases where Fairness is of low importance, and Computational Efficiency is highly valued.

On the other hand, the Proto method is observed to impart a higher degree of personalization to the FL algorithm compared to the MAML method. As shown in Figure \ref{fig:Personalization}, Proto personalization yields a 0.01, 1.11, and 0.78 higher MPI than MALA for FedAvg, FedDyn, and SCAFFOLD, respectively. The enhanced personalization results in higher performance improvements in Computational Efficiency, which is larger than those achieved through the MAML method by approximately 0.3. Conversely, the Proto method introduces a reduction in Fairness, exacerbating the accuracy disparity among clients by about 0.2, while Accuracy and Convergence keep a similar performance. Hence, although greater personalization can enhance Computational Efficiency performance in non-IID scenarios, the trade-offs, particularly in terms of fairness degradation, are also larger.

From the perspective of use cases, we note that only two algorithms, FedAvg\_Proto and FedDyn\_Proto, both PFL algorithms, are rated as Good levels within the IoT scenario. Similarly, in both smartphone and institutional scenarios, algorithms achieving High and Good performance ratings are generally PFL, including FedAvg\_Proto, FedDyn\_Proto, and FedDyn\_MAML. This observation indicates that despite the trade-offs associated with personalization methods, the overall HEM performance is enhanced, benefiting the application of FL algorithms across all three scenarios. As a result, PFL algorithms should be prioritized for selection within these simulated use cases.

However, since the generation of HEM is closely tied to the Importance Vector and its corresponding quantification, further discussion is warranted on whether improvements in convergence can offset decreases in fairness, ultimately contributing to a high HEM for a specific application. This consideration should be explored in conjunction with an investigation into the performance requirements of the deployment use case.

\section{Conclusion and Future work}
In this paper, we outline our proposed HEM index and use three typical use cases to provide a comprehensive evaluation of given FL algorithms in the respective scenario. Specifically, the application scenarios of IoT, Smartphone, and Institution are included as the representative FL use cases. For each scenario, we identify the evaluation metric components and their corresponding importance vectors. The HEM index is generated through a combination of the evaluation metric components and importance vectors. We experimentally assess various FL and PFL algorithms through HEM proposed in different use cases. The experimental results demonstrate that the HEM index can effectively and efficiently evaluate and select the appropriate FL algorithms for diverse scenarios. Currently, we determine the importance of each metric component for each use case based on the findings of literature review and scenario investigation. Moving forward, one future work will be to create a pragmatic benchmarking for the identification of the importance of each metric component for each use case, thereby refining our HEM approach.

\bibliographystyle{IEEEtran} 
\bibliography{Ref}


%




\end{document}